\documentclass[letterpaper, 10 pt, conference]{ieeeconf}  
\IEEEoverridecommandlockouts                              
\usepackage{graphics}
\usepackage{epsfig}     
\usepackage{textcomp}
\usepackage{times}      
\usepackage{amsmath}    
\usepackage{mathtools}  
\usepackage{amssymb}    
\usepackage{url}
\usepackage{commath}
\usepackage{graphicx}
\usepackage[table,xcdraw]{xcolor}

\usepackage{caption}
\usepackage{multirow}

\usepackage{subcaption}
\usepackage{array} 
\usepackage{url}
\usepackage{hyperref}
\hypersetup{colorlinks,allcolors=black}
\usepackage{matlab-prettifier}
\usepackage{siunitx}
\usepackage{xcolor}

\usepackage{listings}
\definecolor{codegreen}{rgb}{0,0.6,0}
\definecolor{codegray}{rgb}{0.5,0.5,0.5}
\definecolor{codepurple}{rgb}{0.58,0,0.82}
\definecolor{backcolour}{rgb}{0.95,0.95,0.92}

\lstdefinestyle{mystyle}{
    backgroundcolor=\color{backcolour},   
    commentstyle=\color{codegreen},
    keywordstyle=\color{magenta},
    numberstyle=\tiny\color{codegray},
    stringstyle=\color{codepurple},
    basicstyle=\ttfamily\footnotesize,
    breakatwhitespace=false,         
    breaklines=true,                 
    captionpos=t,                    
    keepspaces=true,                 
    numbers=left,                    
    numbersep=5pt,                  
    showspaces=false,                
    showstringspaces=false,
    showtabs=false,                  
    tabsize=2,
    frame=single
}

\DeclareCaptionFormat{listing}{\rule{\dimexpr\textwidth+17pt\relax}{0.4pt}\par\vskip1pt#1#2#3}

\title{\LARGE \bf
SMARTmBOT: A ROS2-based Low-cost and 
\\Open-source Mobile Robot Platform}

\author{Wonse Jo, Jaeeun Kim, Ruiqi Wang, Jeremy Pan, Revanth Krishna Senthilkumaran, and Byung-Cheol Min
\thanks{All authors are with SMART Lab, Department of Computer and Information Technology, Purdue University, West Lafayette, IN 47907, USA \tt\small{[jow, kim2592, wang5357, pan117, senthilr, and minb] @purdue.edu}}%
}

\begin{document}
\maketitle
\thispagestyle{empty}
\pagestyle{empty}

\begin{abstract} 
This paper introduces SMARTmBOT, an open-source mobile robot platform based on Robot Operating System 2 (ROS2). The characteristics of the SMARTmBOT, including low-cost, modular-typed, customizable and expandable design, make it an easily achievable and effective robot platform to support broad robotics research and education involving either single-robot or multi-robot systems. The total cost per robot is approximately \$210, and most hardware components can be fabricated by a generic 3D printer, hence allowing users to build the robots or replace any broken parts conveniently. The SMARTmBot is also equipped with a rich range of sensors, making it competent for general task scenarios, such as point-to-point navigation and obstacle avoidance. We validated the mobility and function of SMARTmBOT through various robot navigation experiments and applications with tasks including go-to-goal, pure-pursuit, line following, and swarming. 
All source code necessary for reading sensors, streaming from an embedded camera, and controlling the robot including robot navigation controllers is available through an online repository that can be found at \url{https://github.com/SMARTlab-Purdue/SMARTmBOT}.
\end{abstract}

\section{Introduction}
\label{sec:introduction}

Development and innovation in the area of mobile robot platforms has been underway since the 1990s in the interest of continually integrating up-to-date sensors and software \cite{siegwart2011introduction}. In the early stages of mobile robot development, microprocessors constituted fundamental resources for robot control and reading sensor data \cite{tzafestas1991microprocessors}. However, with the emergence of artificial intelligence (AI) technology, mobile robots have pivoted to use single-board computers (SBCs) that have higher computing power and so support more advanced AI technology and sensor modules \cite{norris2017beginning}. Simultaneously, Robot Operating System (ROS) \cite{quigley2009ros} has significantly accelerated the development of SBC-based mobile robots. Nonetheless, developing such robots from scratch remains expensive for researchers, and the process is time-consuming. An alternative is to use existing mobile robot platforms as research tools, but many of the currently-available platforms have only limited expandability and compatibility with additional hardware, and moreover cannot connect with other systems due to a lack of unified communication protocols.

\begin{figure}[t]
    \centering
        \includegraphics[width=0.9\linewidth]{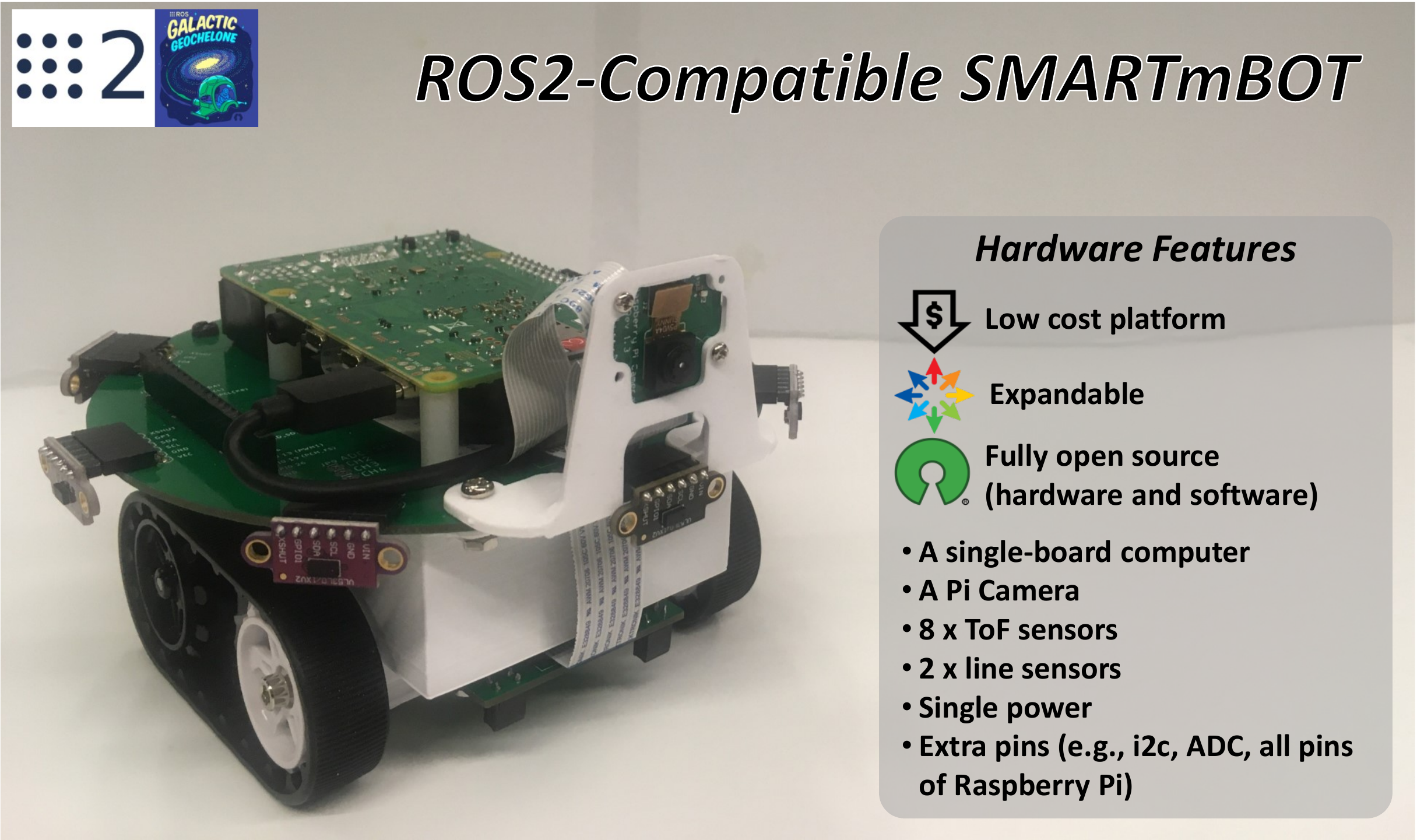}  
    \caption{Key features of the proposed mobile robot platform, SMARTmBOT, which is fully open-source, low-cost, modular-type, customizable, and expandable.} 
    \label{img:3dmodel}
\end{figure}

For this purpose, we propose a new, fully open-source, ROS2-based low-cost mobile robot platform called the \textbf{SMART} lab \textbf{m}obile-ro\textbf{BOT} (SMARTmBOT) that can be used in research and educational settings. The SMARTmBOT is made with a 3D-printed casing and runs Robot Operating System 2 (ROS2) \cite{ros_2_overview} to control the components and read the sensors mounted on modular-type printed circuit board (PCB) layers. It features eight time-of-flight (ToF) laser-ranging sensors, two infrared phototransistors (line sensors), and two DC geared motors with a motor driver. The robot is powered by a portable power bank, which serves as a unified power source to maximize convenience. Key features of the SMARTmBOT are depicted in Fig. \ref{img:3dmodel}. We have validated the mobility, expandability, and potential applications of the SMARTmBOT platform through sensor and robot navigation experiments. We have also created an online repository on GitHub to share all source code and mechanical and electronic design files to ensure the easy use and dissemination of the SMARTmBOT platform.

\section{Related works}
\label{sec:liturature_reviews}
A wide variety of mobile robot platforms have been introduced in which advanced robotic technologies are combined, such as sensors, Lidars, and middleware packages. Existing platforms can be divided into two categories according to processor types. 

One main category of robot platforms is those having a microprocessor unit (MCU) as the processor. Using a microprocessor allows the robot to be lighter, cheaper, and smaller, with more freedom in designing the circuit. However, it might be difficult for entry-level programmers to fully utilize its potential.  
Kilobot \cite{kilobot} is a well-known platform in this category due to its low-cost and reliable performance, but it is not customizable due to limited space on its hardware structure. GRITSbot \cite{gritsbot} is another example that is small and low-cost with customizable hardware, but does not support ROS or ROS 2. E-puck2 and Elisa are similar MCU-based platforms that provide small, lightweight, expandable devices for multi-robot research \cite{epuck}. However, those robots start at \$500, which makes them unsuitable for large-scale multi-robot systems under a limited budget. Turtlebot \cite{turtlebot} has several options for customization and is officially recognized and well-supported in the ROS community, but is limited in application by its relatively large size and can be cost-prohibitive in a research setting, having a retail price starting about \$600. iRobot's Roomba \cite{roomba} is often utilized in academia for robotics projects; however, since that platform was originally developed as a vacuum robot for home use, it not only has limited flexibility when it comes to altering software or hardware, but also requires additional third-party software that has no guarantees for full utilization of the platform's capabilities.

The Zumo \cite{zumo} is an Arduino-compatible integrated robot. 
While the simplicity offered by the Arduino has the benefit of being more accessible to a larger audience, this platform has limited computing power, and ROS2 integration requires potentially difficult-to-obtain additional components (e.g., adding a SBC). R-One \cite{mclurkin2013low} is a microprocessor-based platform that features a camera and infrared (IR) receivers and transmitters in order to enable inter-robot communication and localization. A grip module \cite{mclurkin2014robot} is also available, allowing it to perform multi-robot manipulation. However, the R-One has a complex hardware design, is unaffordable for beginners, and seems to have been discontinued. Finally, the Khepera IV \cite{khepera} is a commercial platform targeted at education and research tasks. It has a range of basic sensors and an additional manipulator module, and is expandable for use as a multi-robot system. However, the price of a Khepera robot with all modules is about \$6,000, making the platform unaffordable.

Another main type of robot platforms is those using a single-board computer (SBC) as the processor. SBCs are well-supported both officially and unofficially in online forums. They may be beginner friendly, but can be costly, especially when bought in large quantities. 
However, there is only one open-source platform available in this category, the recently-launched JetBot \cite{jetsonbot}. This platform features a new SBC including a GPU, but the ecosystem of the Nvidia Jetson is still smaller than that for the Raspberry Pi, which can lead to difficulty for users learning robotics.  

\begin{table}[t]
    \caption{Comparison of existing mobile robot platforms and the SMARTmBOT} 
    \label{tab:compare}
    \resizebox{\columnwidth}{!}{%
    \begin{tabular}{ |c|c|c|c|c|c|c|} 
        \hline
        \rowcolor{gray}
        \textbf{Name} & \textbf{ROS} & \textbf{SBC} & \textbf{MCU}  & \textbf{Custom}  & \textbf{Open}  & \textbf{Price}\\ 
        \hline
        \hline
        Kilobot & & & O & & & \~\$ 120 \\  
        \hline
        GRITSbot & & & O & & & \~\$ 50 \\  
        \hline
        E-puck & O & & O & O & & \~\$ 930 \\  
        \hline
        Elisa & & & O & O & & \~\$ 350 \\  
        \hline
        Turtlebot & O & & O & O & O & \~\$ 549 \\  
        \hline
        iRobot &  & & O & & & \~\$ 130 \\  
        \hline
        Zumo & O & & O & O & & \~\$ 100 \\  
        \hline
        R-One & & & O & & O & \~\$ 250 \\  
        \hline
        Khepera IV & & & O & & O & \~\$ 2900 \\  
        \hline
        JetBot & O & O$^{N}$ & & O &  & \~\$ 250 \\  
        \hline
        SMARTmBot & O & O$^{R}$ & & O & O & \~\$ 210 \\  
        \hline
    \end{tabular}
    }
    \footnotesize{$^{N}$: Nvidia Jetson Nano, $^{R}$: Raspberry Pi 4.}
\end{table}

Table \ref{tab:compare} summaries all existing platforms that are similar to ours, where ROS indicates whether the platform supports ROS or not; SBC and MCU whether the platform is mainly operated by a single-board computer (SBC) or microprocessor (MCU), respectively; Custom whether the platform supports easy modification of its design and adding of sensors; and Open whether both the hardware (electronic circuit, design files, etc.) and software (low-level source, ROS2 packages, etc.) are fully open-source.

All told, while diverse existing mobile robot platforms that share similar funcitons are available, each has substantial limitations such as unaffordable costs and limited compatibility with other systems. Thus, we propose a new ROS2-based, fully open source, low-cost mobile robot platform, called SMARTmBOT, that supports up-to-date ROS and can be easily combined with other robotics systems by beginners in the course of their research.





\section{Mobile robot platform: SMARTmBOT}
\subsection{Overview of platforms}
\label{sec:hardware}
The proposed SMARTmBOT supports ROS2 and is composed of off-the-shelf components as shown in Table \ref{tab:bom}, to better assist those who are building the robot for personal and research purposes. The cost of building a SMARTmBOT is roughly \$210, which is much more affordable than other existing platforms. The robot is 15 $cm$ in diameter and 10 $cm$ in height, 900 $grams$ in weight, and 90\% of its components are 3D-printed parts. All of the necessary materials, including the 3D design CAD files, PCB design files, and ROS2 packages, are fully open-source and can be found at the following GitHub repository: \url{https://github.com/SMARTlab-Purdue/SMARTmBOT}.

\begin{table}[t]
    \centering
    \caption{A bill of materials (BOM) list for building a SMARTmBOT} 
    \label{tab:bom}
    \resizebox{\columnwidth}{!}{%

    \begin{tabular}{|c|l|c|c|} 
        \hline
        \rowcolor{gray}
        \textbf{\#} & \textbf{\centering Item} & \textbf{Quantity} & \textbf{Price}\\ 
        \hline
        \hline

        1 & Power Bank & 1 & \$ 22.99 \\  
        \hline
        2 & DC Gear Motor & 2 & \$ 17.95 \\  
        \hline
        3 & 32GB SD Card & 1 & \$ 7.49 \\  
        \hline
        4 & Raspberry Pi 4  & 1 & \$ 35.0$~$ \\  
        \hline
        5 & Pi Camera & 1 & \$ 7.99 \\  
        \hline
        6 & 22T Track Wheel Set & 1 & \$ 11.66 \\  
        \hline
        7 & RPR220 Photo Reflector Sensor & 2 & \$ 1.10 \\  
        \hline
        8 & \begin{tabular}[l]{@{}l@{}}VL53L0X Time-of-Flight \\Laser Ranging Sensor \end{tabular} & 8 & \$ 7.45 \\  
        \hline
        9 & 550 SMD RGB LED & 4 & \$ 0.53 \\  
        \hline
        10 & MCP3008 & 1 & \$ 1.82 \\  
        \hline
        11 & L293D & 1 & \$ 3.32 \\  
        \hline
        12 & \begin{tabular}[l]{@{}l@{}}2.5mm JST Female \\and Male Connector \end{tabular}  & 3 sets & \$ 0.45 \\  
        \hline
        13 & FFC Connector & 2 & \$ 1.28 \\  
        \hline
        14 & FFC Cable & 1 & \$ 2.44 \\  
        \hline
        15 & \begin{tabular}[l]{@{}l@{}}100 Ohm \& 200 Ohm \\ \& 68k Ohm Resistor \end{tabular} & 1 & \$ 0.11 \\  
        \hline
        16 & 90 Degree Angle Pin Header & 8 & \$ 0.80 \\  
        \hline
        17 & Long 20 Pin Female Header & 2 & \$ 1.37 \\  
        \hline
        18 & 20 Pin Female and Male Header & 2 sets & \$ 0.34 \\  
        \hline
        19 & Spacer (5mm \& 100mm) & 4 sets & \$ 0.72 \\  
        \hline
        20 & M3 Bolts and Nuts & 20 sets & \$ 0.2 \\  
        \hline
    \end{tabular}
    }

\end{table}

\subsection{Hardware Design of the Modular-type Platform}
Fig. \ref{img:robot_hardware_parts} depicts all of the components used in the proposed SMARTmBOT, including 3D-printed parts and PCB layers. There are four 3D-printed parts that can be modified by users: the battery holder, two motor brackets, shaft support brackets, and transparent case. The battery holder serves as the main base of the SMARTmBOT. The two motor mount brackets each hold a DC motor and are attached to the power bank holder. The shaft support brackets likewise secure the two rear wheels to the power bank holder. The front and rear wheels are connected with a rubber chain belt to improve the stability of the platform. The transparent case is to cover and protect the entire platform.

In terms of the PCB layers, there are three layers that are connected with 3D-printed parts to build a robot platform. 
\subsubsection{First PCB Layer}
The first PCB layer, as shown in Fig. \ref{img:line_layer}, is located at the bottom and between the wheels, facing downwards. It includes two DC geared motors with an L293D motor driver, two photo reflector sensors, and an MCP3008 8-channel 10-bit analogue-to-digital converter (ADC) IC chip, which converts ADC values to digital data and sends it through serial peripheral interface (SPI) communication. 
The DC motors used in the SMARTmBOT are a miniature high-power, 12 V DC motor with carbon brushes and a 100:1 metal gearbox. The no-load speed is 330 RPM, the extrapolated stall torque is 1.3 kg$\cdot$cm, and the stall current is 0.75 A. The motors are controlled by a L293D motor driver which provides bidirectional drive currents of up to 1 A at voltages between 4.5 V and 36 V.
Motor control lines and external power supplied from the power bank are also connected to this layer.
For feeding the DC motors, a DC step-up module was mounted on the SMARTmBOT, which also helps users easily replace different motors by adjusting the voltages of the step-up module range from 4.5V to 36V. 

\subsubsection{Second PCB Layer}
The second layer is placed on top of the battery holder, holding mounts for the Raspberry Pi, a Pi camera module, and the eight ToF sensors surrounding the platform with 45 degrees of separation between each, as shown in Fig. \ref{img:tof_layer}. 
The Pi camera module is the Raspberry Pi HD Camera Module V2 that has an 8 megapixel high-quality camera. 
The ToF is a VL53L0X laser-ranging sensor that can detect objects of up to a maximum of 2 meters away. Both the first and the second PCB layers are connected via flat flexible cables (FFC). 
\subsubsection{Third PCB Layer}
\begin{figure}[t]
    \centering
     \begin{subfigure}[b]{0.52\linewidth}
        \centering 
        \begin{subfigure}[b]{1\linewidth}
            \centering 
            \includegraphics[width=1\linewidth]{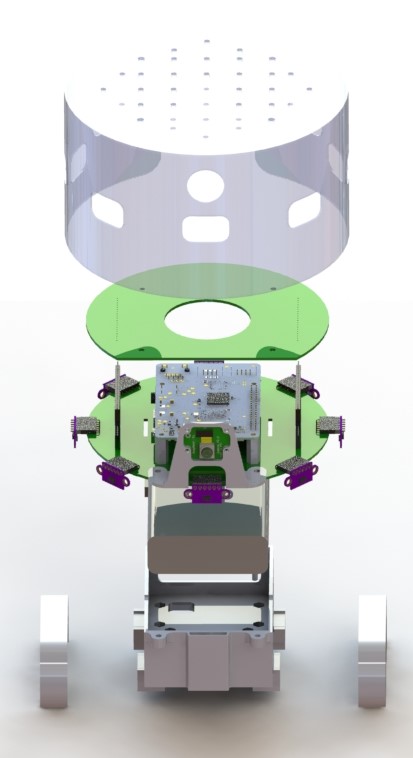} 
            \caption{}
            \label{img:robot_parts}
        \end{subfigure}
        
        \begin{subfigure}[b]{0.95\linewidth}
            \centering 
            \includegraphics[width=1\linewidth]{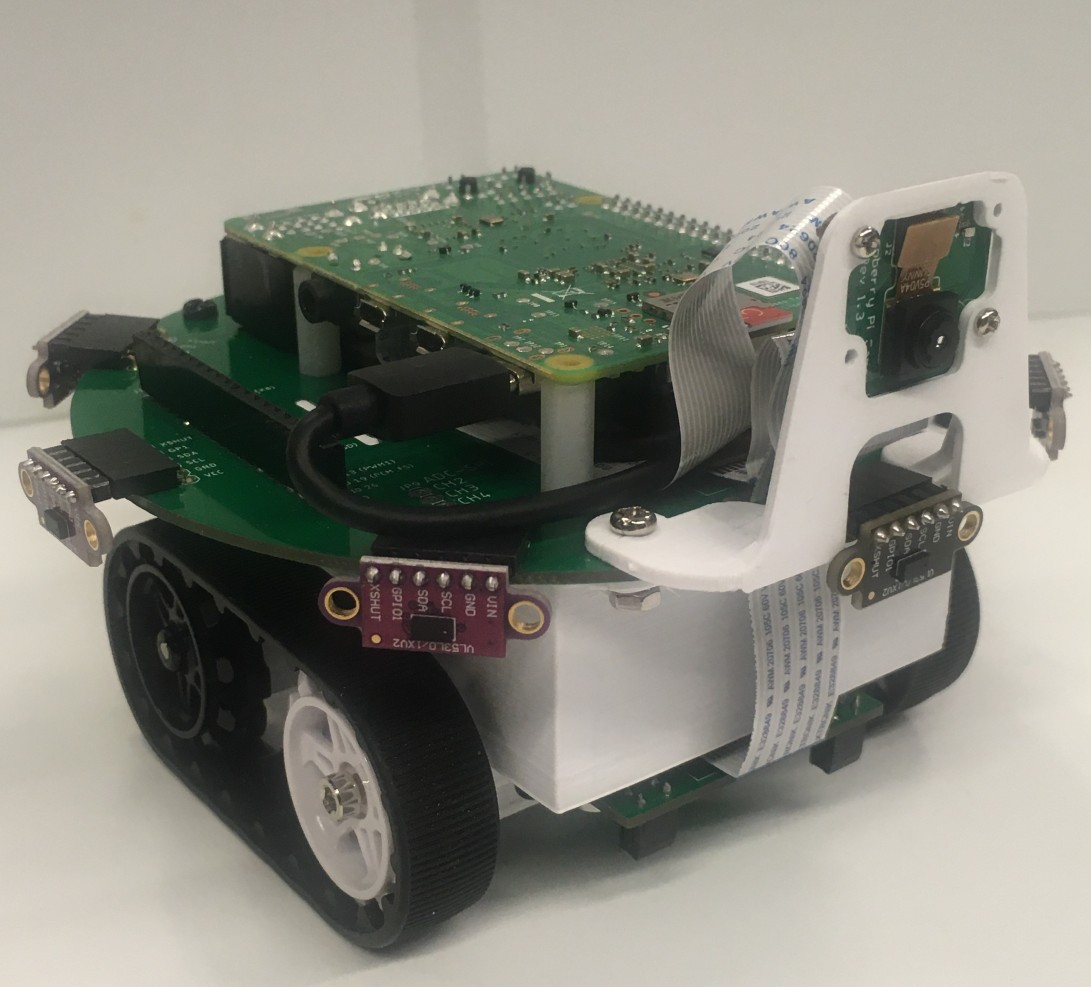} 
            \caption{}
            \label{img:3d_hardware_part}
        \end{subfigure}

    \end{subfigure}
    \begin{subfigure}[b]{0.45\linewidth}
        \begin{subfigure}[b]{1\linewidth}
            \centering 
            \includegraphics[width=1\linewidth]{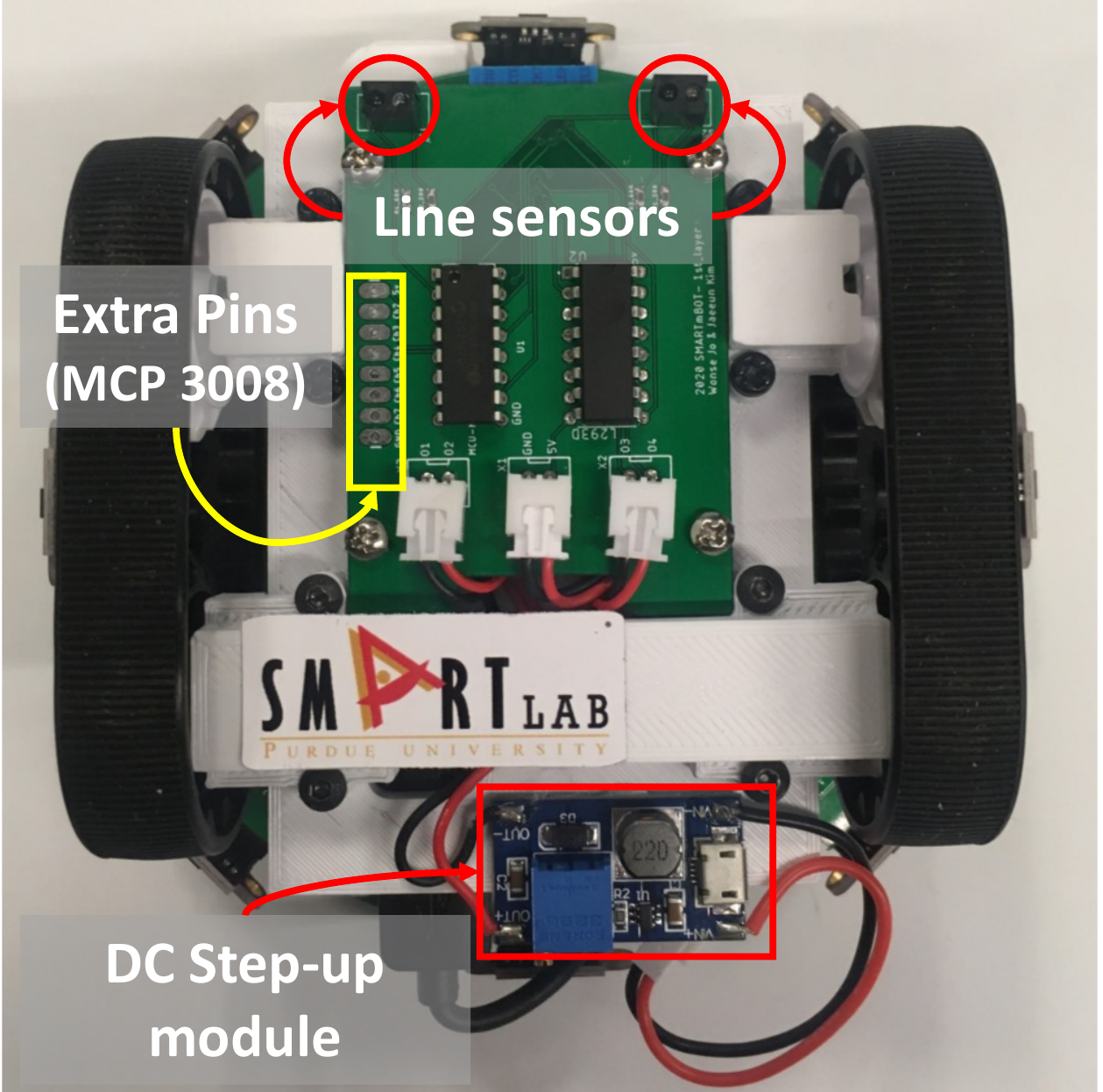} 
            \caption{}
            \label{img:line_layer}
        \end{subfigure}
        
        \begin{subfigure}[b]{1\linewidth}
            \centering             \includegraphics[width=1\linewidth]{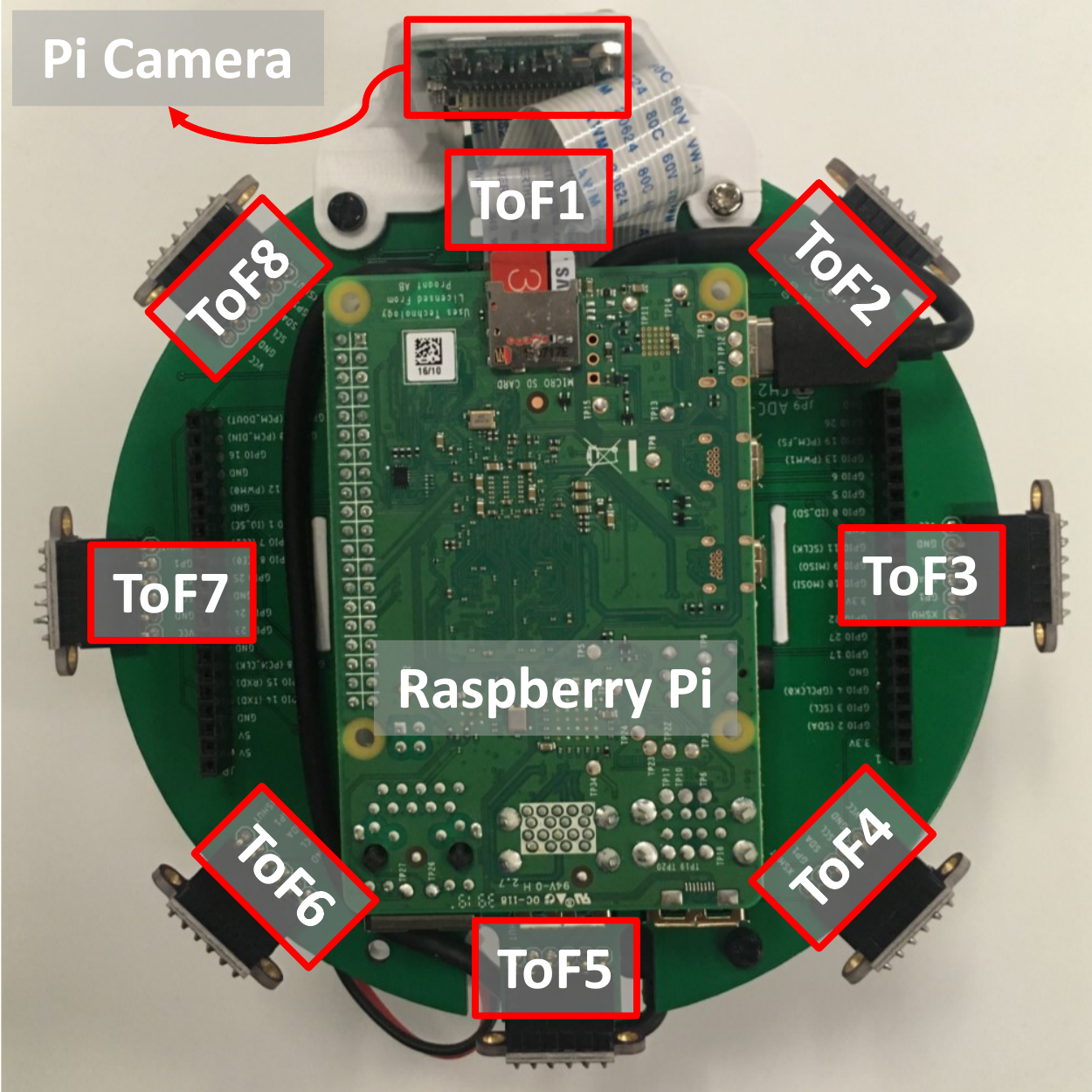} 
            \caption{}
            \label{img:tof_layer}
        \end{subfigure}
        
         \begin{subfigure}[b]{1\linewidth}
            \centering             \includegraphics[width=1\linewidth]{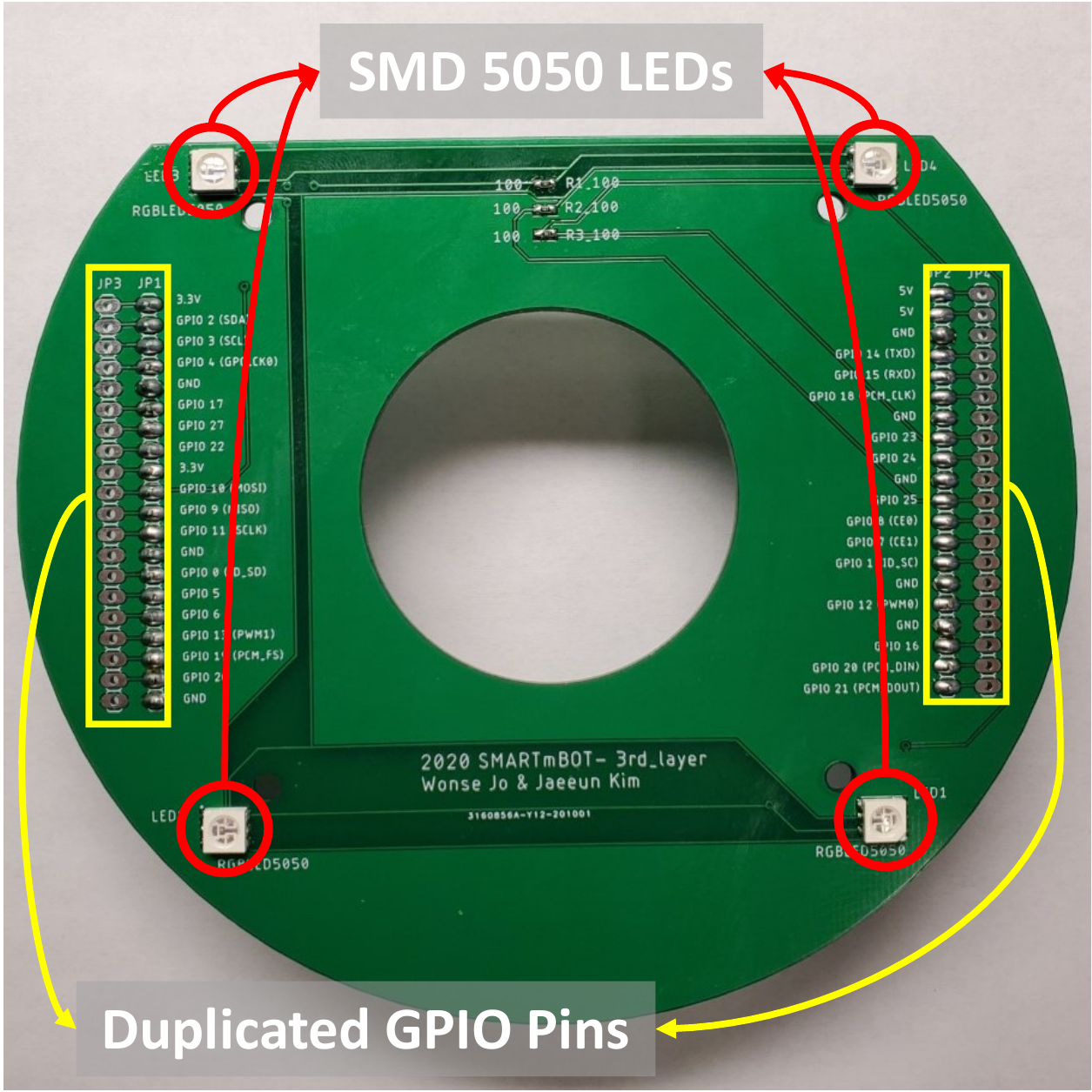} 
            \caption{}
            \label{img:third_layer}
        \end{subfigure}
    \end{subfigure}
    
    \caption{Hardware structures of SMARTmBOT: (a) exploded view drawing of the platform, (b) fully assembled SMARTmBOT (without a case), (c) first PCB layer mounted on bottom of the platform that includes line sensors, a DC motor driver, and a SPI-ADC chip, (d) second PCB layer including eight ToF sensors, Raspberry Pi, and a Pi camera, and (e) third PCB layer including SMD-type LEDs and duplicatfor the paper, titleded GPIO pins.}
    \label{img:robot_hardware_parts}
\end{figure}

The third PCB layer is placed on top of the second PCB layer and has four SMD 5050 LEDs capable of changing their colors and intensities, as shown in Fig \ref{img:third_layer}. It also has a pillafor the paper, titledr role between the second PCB layer and the case to make the platform stable. The third layer can be replaced with other PCBs or 2.54 $mm$ pitch solder-in breadboards since it has a unified 2.54 $mm$ distance between holes. The second and third layers are connected and shared with all GPIO pins of the SBC via pin headers with the 2.54 $mm$ pitch and provide users with easy access to all functions through the duplicated GPIO pins.

\subsection{Design of ROS2 Software Architecture}
\label{sec:software}

\begin{figure*}[t]
    \centering
        \includegraphics[width=1\linewidth]{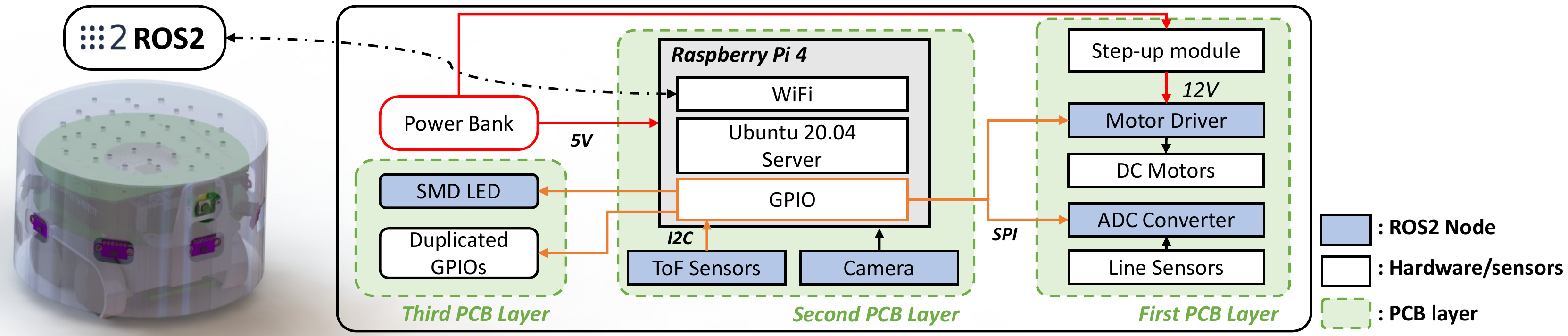}
    \caption{A diagram of the software architecture used in the SMARTmBOT.}
    \label{img:software_part}
\end{figure*} 

\begin{figure}[t]
    \centering             
    \includegraphics[width=1\linewidth]{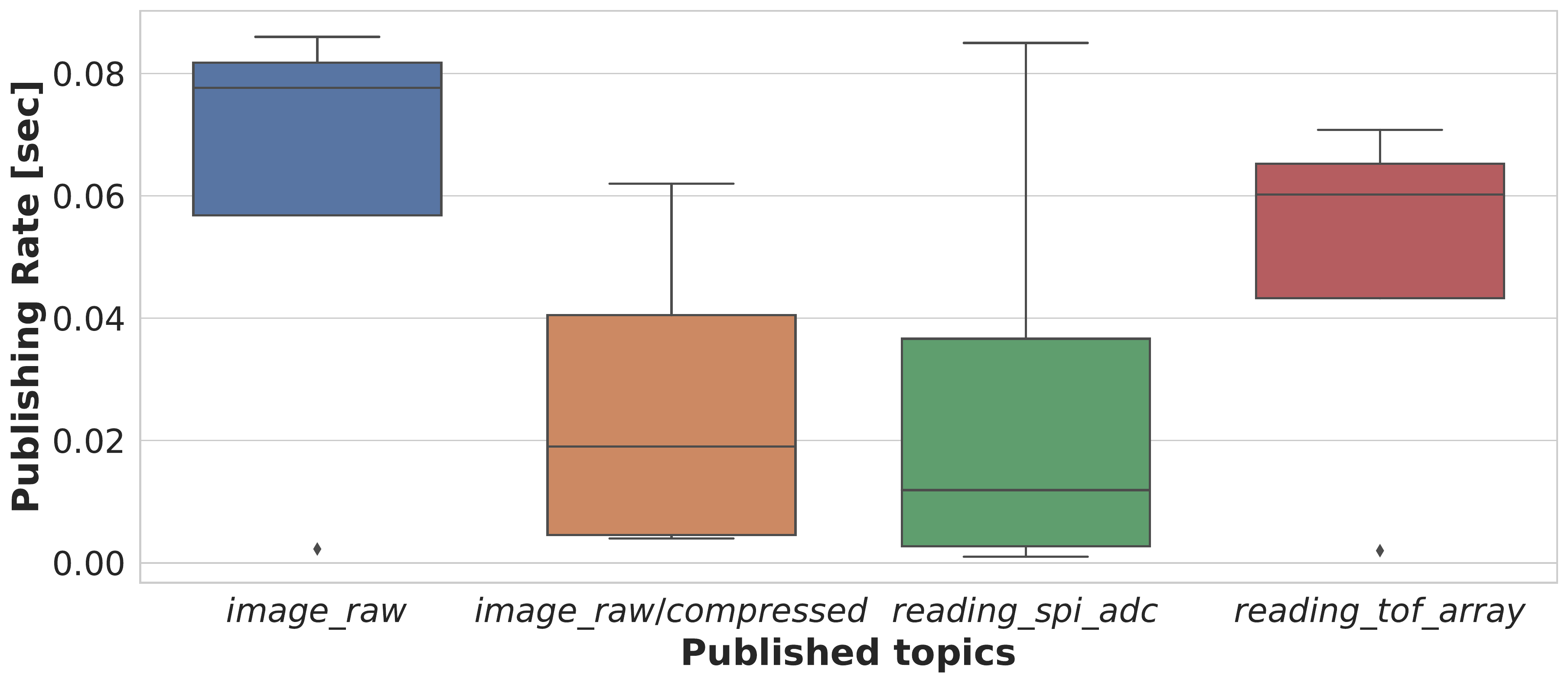}
    \caption{Average sampling rate of published topics with standard deviation: raw image, compressed image, SPI-based eight ADC values, and eight ToF sensors (from left to right).} 
    \label{img:topic_rate}
\end{figure}

Fig. \ref{img:software_part} illustrates the overall software architecture used in the SMARTmBOT ROS package, which was developed with Python 3.8 in Ubuntu 20.04. The Raspberry Pi 4 acts as the main processing computer, which runs on Ubuntu 20.04. With its GPIO pins, the Raspberry Pi controls the DC motors via a motor driver by means of adjusting pulse-width modulation (PWM), reads ToF sensors over inter-integrated circuit (I2C) communication, reads analog infrared phototransistors with an ADC converter over SPI communication, and controls the color of RGB LEDs. Each control function is built as a separate node in ROS2, as listed in Table \ref{tab:ros2_topic_list}. 
To validate the sampling rate of each topic, we read the topic using a remote machine that was connected to the ROS2 system. The average of the measured sampling rate of each published topic on the remote machine is shown in Fig. \ref{img:topic_rate}.

\begin{table*}[]
    \caption{ROS2 Topic List of SMARTmBOT}
    \label{tab:ros2_topic_list}
    \resizebox{\textwidth}{!}{%
    \begin{tabular}{|c|l|l|c|l|}
        \hline
        \rowcolor[HTML]{C0C0C0} 
        \textbf{Type} & \multicolumn{1}{c|}{\cellcolor[HTML]{C0C0C0}\textbf{Name}} & \multicolumn{1}{c|}{\cellcolor[HTML]{C0C0C0}\textbf{Data type}} & \textbf{\begin{tabular}[c]{@{}c@{}}Sampling\\ rate \end{tabular}} & \multicolumn{1}{c|}{\cellcolor[HTML]{C0C0C0}\textbf{Description}} \\ \hline
        Pub. & /\{robot name\}/image raw & sensor\_msgs/msg/Image & 12 Hz& \begin{tabular}[l]{@{}l@{}}Camera views streamed from SMARTmBOT \end{tabular} \\ \hline
        Pub. & /\{robot name\}/image\_raw/compressed & sensor\_msgs/msg/CompressedImage & 30 Hz& \begin{tabular}[l]{@{}l@{}}Compressed camera views streamed\\ from SMARTmBOT\end{tabular} \\ \hline
        Pub. & /for the paper, titled\{robot\_name\}/reading\_spi\_adc & std\_msgs/msg/Float32MultiArray & 50 Hz& Array of eight SPI sensor values. \\ \hline
        Pub. & /\{robot\_name\}/reading\_tof\_array & std\_msgs/msg/Float32MultiArray & 15 Hz& Array of eight ToF sensor values. \\ \hline
        Sub. & /\{robot\_name\}/writing\_dc\_cmd\_vel & geometry\_msgs/msg/Twist & n/a & \begin{tabular}[l]{@{}l@{}}Linear and angular velocities of SMARTmBOT \\  \end{tabular} \\ \hline
        Sub. & /\{robot\_name\}/writing\_dc\_motor\_vel & std\_msgs/msg/Float32MultiArray & n/a & PWM control for DC motors \\ \hline
        Sub. & /\{robot\_name\}/writing\_gpio\_smd5050\_led & std\_msgs/msg/Float32MultiArray & n/a & \begin{tabular}[l]{@{}l@{}}Control for color and intensity of SMD5050 \\mounted on the third PCB layer \end{tabular} \\ \hline
    \end{tabular}%
    }
\end{table*}

\subsubsection{Pi Camera Node}
This node streams the two types of the images recorded from the Pi camera: raw images and compressed images. The topic names are \textit{image\_raw} and  \textit{image\_raw/compressed} with sampling rate of 12 Hz and 30 Hz, respectively. The quality of the raw image is better than the compressed image, but it has more latency than the compressed image.  

\subsubsection{Reading SPI ADC Node} 
This node publishes the converted 8-channel ADC values from the MCP3008 that is mounted on the first PCB layer of the SMARTmBOT. 
The published topic name in this node is \textit{/[robot\_name]/reading\_spi\_adc} with sampling rate of 50 Hz. The topic message type is Float32MultiArray standard message. The $data[0]$ and $data[1]$ of the topic message are the left and right infrared phototransistors, respectively. The other data from $data[2]$ to $data[7]$ are ADC values of the remaining pins of the MCP3008 so the user can easily add additional analog sensors. 

\subsubsection{Reading I2C ToF Node}
This node publishes the measured distances in millimeters of the eight VL53L0X ToF sensors mounted on the second PCB layer. 
The published topic name of this node is \textit{reading\_tof\_array}, which has a sampling rate of 15 Hz. The topic message type is a Float32MultiArray standard message. The distances measured by each ToF sensor is represented in fields $data[0]$ to $data[7]$, listed in clockwise order from the head of the robot.

\subsubsection{Writing DC Motor Node}
This node controls two DC motors using two types of ROS2 topic messages: Float32MultiArray and Twist. Each motor's speeds can be altered separately using a PWM signal from the Raspberry Pi. The topic names are \textit{writing\_dc\_motor\_vel} and  \textit{writing\_dc\_cmd\_vel}. 
The data of the Float32MultiArray-type ROS2 topic is PWM values for the two DC motors ranging from -100 to 100. 
The data of the Twist-type ROS2 topic includes both linear $v$ and angular velocities $\omega$. Then, the data are automatically converted into PWM values in the Writing DC Motor Node using a differential drive kinematics \cite{kim20041}:
\begin{align} \label{eq:differential_dirve}
    \begin{split}
        v = \frac{V_{R} + V_{L}}{2} = r(\frac{\omega_{R}+\omega_{L}}{2})
        \\
        \omega = \frac{V_{R} - V_{L}}{w_{b}}= r(\frac{\omega_{R}-\omega_{L}}{w_{b}})
    \end{split}
\end{align}
where $V_{R}$ and $V_{L}$ are the linear velocities of each wheel, $\omega_{R}$ and $\omega_{L}$ are the angular velocity of each wheel, $v$ is the linear velocity and $\omega$ is the angular velocity of the robot, $r$ is the radius of the wheel, and $w_{b}$ is the distance between the left and right wheels.

\subsubsection{Writing GPIO SMD5050 LED Node}
This node controls the four SMD type 5050 RGB LEDs mounted on the third layer of the PCB board. The topic name is \textit{writing\_gpio\_smd5050\_led} that provides Float32MultiArray data. It can adjust the color and intensity of the RGB LEDs. 

\section{Function and Performance Testing}
\label{sec:function_test}
To verify the capabilities of the SMARTmBOT, we have conducted real-world experiments and obtained measurable results regarding system's performance. We gathered data from the sensors and validated the differential drive mechanism of the platform through the following tests. A video that shows each test is available on our GitHub repository.

\subsection{Sensor Testing}
To validate the functions of the sensor systems, we conducted two experiments: object detection using the eight ToF sensors and line tracing using two infrared phototransistors. 

\subsubsection{Measuring distances using eight ToF sensors} 
Fig. \ref{img:sensor_exp_1} depicts the result of measuring distances of objects using eight ToF sensors. There were four objects (i.e., other robots) placed roughly 200 $mm$ away from the SMARTmBOT. The topic data read by the I2C ToF node were 
listed in clockwise order from the head of the robot. By doing so, the SMARTmBOT could detect the four objects simultaneously with a sampling rate of 15 Hz.

\subsubsection{Line tracking algorithm using two infrared phototransistors} 
Fig. \ref{img:sensor_exp_2} depicts the trajectory of the SMARTmBOT while following the black line drawn on the floor by using two infrared phototransistors. The sensors were mounted on the first PCB layer facing downward. The wheel speeds were changed based on the ADC values of the infrared phototransistors. If the left sensor is higher than the threshold value, the speed of the right wheel increased for the SMARTmBOT to rotate to the right. By repeating these steps, the SMARTmBOT could follow the line and arrive at the goal position.


\subsection{Mobility Testing}
To validate the mobility of the SMARTmBOT, we experimented with two different robot navigation algorithms: the go-to-goal algorithm and the pure-pursuit algorithm. Through a motion capture system, we measured and tracked the global position and orientation of the SMARTmBOT. Both robot navigation algorithm nodes (go-to-goal node and pure-pursuit node) generated wheel velocities individually to move to a goal position with a desired heading angle. In addition, in order to validate the reproducibility of the SMARTmBOT, multiple robots were manufactured using the same manufacturing process and components, and the two navigation algorithm tests were conducted. 

\subsubsection{Go-to-goal algorithm} The algorithm is a proportional controller based on the two-differential drive kinematics model \cite{gotogoal}. For this experiment, a goal position was pre-defined ($x^{*}, y^{*}$) at the center of the four robots ($x^{s}_{i}, y^{s}_{i}$), $i = {1, ... , 4 }$. Then, the go-to-goal algorithm calculated the desired velocity and heading angle of each robot as follows:
\begin{align} \label{eq:go_to_goal}
    \begin{split}
        v_{i} = K^{linear}_{p} * \sqrt{(x^{*}-x^{s}_{i})^2 + (y^{*}-y^{s}_{i})^2} \\
        \theta_{i} = K^{angular}_{p}( arctan\frac{y^{*}-y^{s}_{i}}{x^{*}-x^{s}_{i}} - \theta^{c})
    \end{split}
\end{align}
\noindent where $\theta$ is the goal angle, $Kp$ is the proportional gain, and $\theta^{c}$ is the current angle.

As a result, all four SMARTmBOTs (R1, R2, R3, and R4) could arrive at their goal positions. Their trajectories are illustrated in Fig. \ref{img:go_to_goal_exp}, where the colored dots indicate the starting positions of each robot while the star markers indicate the goal positions.

\subsubsection{Pure-pursuit algorithm}

\begin{figure}[t]
    \centering             
    \includegraphics[width=0.7\linewidth]{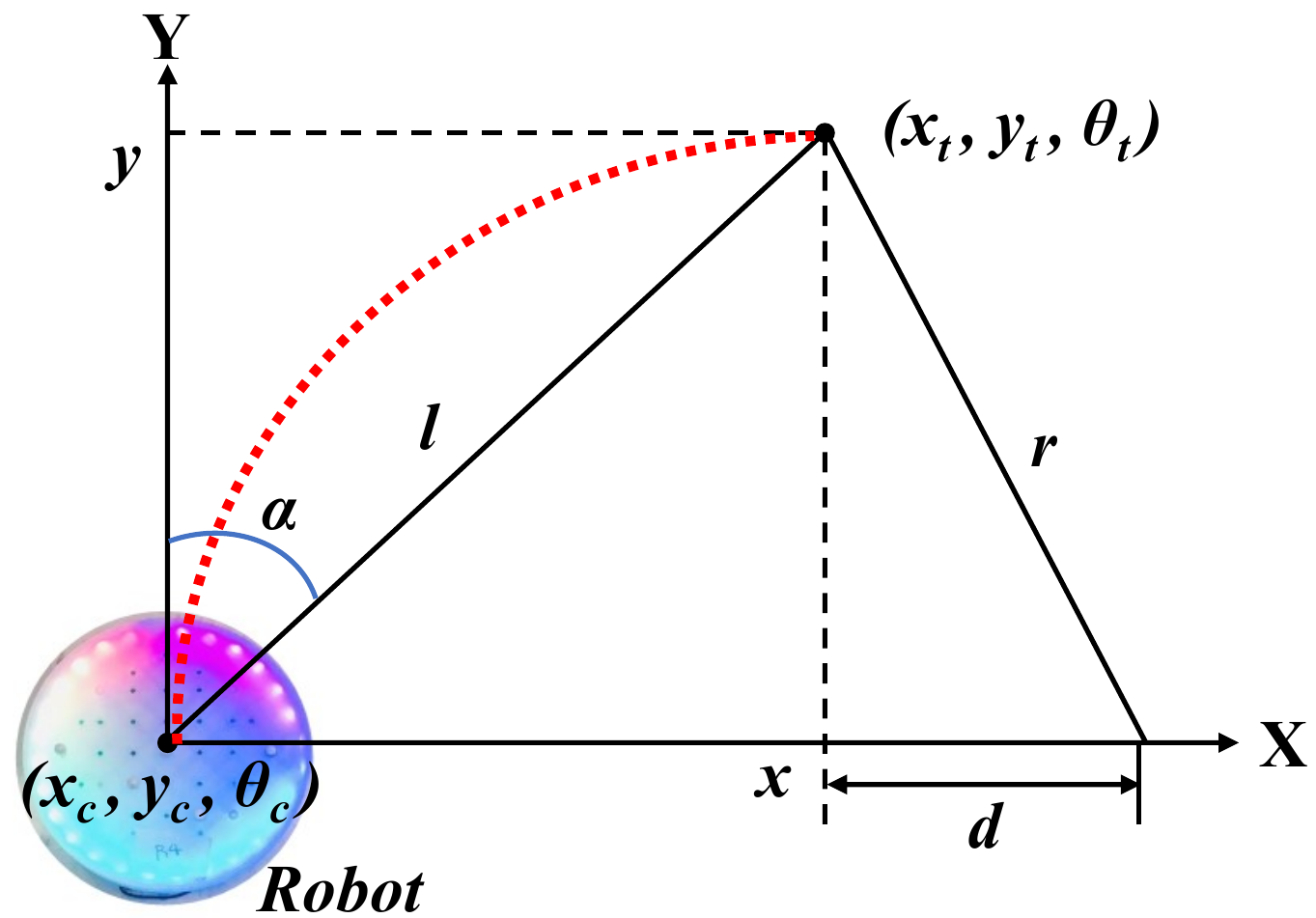}
    \caption{Geometry of the pure pursuit algorithm.} 
    \label{img:pure_pursuit_model}
\end{figure}
This algorithm is a popular navigation algorithm for autonomous vehicles to track a path by steering a heading angle, called a curvature ($\delta$). The multiple waypoints in the path are calculated from a starting point to a goal point. 
It drives to a look-ahead point $(x_{t}, y_{t})$ by adjusting linear and angular velocities using the $\delta$ as follows \cite{paden2016survey}: 
\begin{align} \label{eq:pure_pursuit}
    \begin{split}
        \alpha = K_{p}^{angular}\tan^{-1}(\frac{x_{t} - x_{c}}{y_{t} - y_{c}}) - \theta_{c} \\
        \delta  = \tan^{-1}(\frac{K_{p}^{linear}w_{b} \sin \alpha }{l})
    \end{split}
\end{align}

\noindent where $K_{p}^{angular}$ and $K_{p}^{linear}$ are a proportion gain for the angular velocity and linear velocity, respectively, ($x_{c}$, $y_{c}$) is a current robot position, ($x_{t}$, $y_{t}$) is a look-ahead point, $\alpha$ is the error angle between ($x_{c}$, $y_{c}$) and ($x_{t}$, $y_{c}$), $\theta_{c}$ is a current heading angle, and $l$ is a look-ahead distance between the look-ahead point and the current robot position, as illustrated in Fig.~\ref{img:pure_pursuit_model}.

\begin{figure*}
    \centering
    \begin{subfigure}[b]{0.20\linewidth}
        \centering
        \includegraphics[width=1\linewidth]{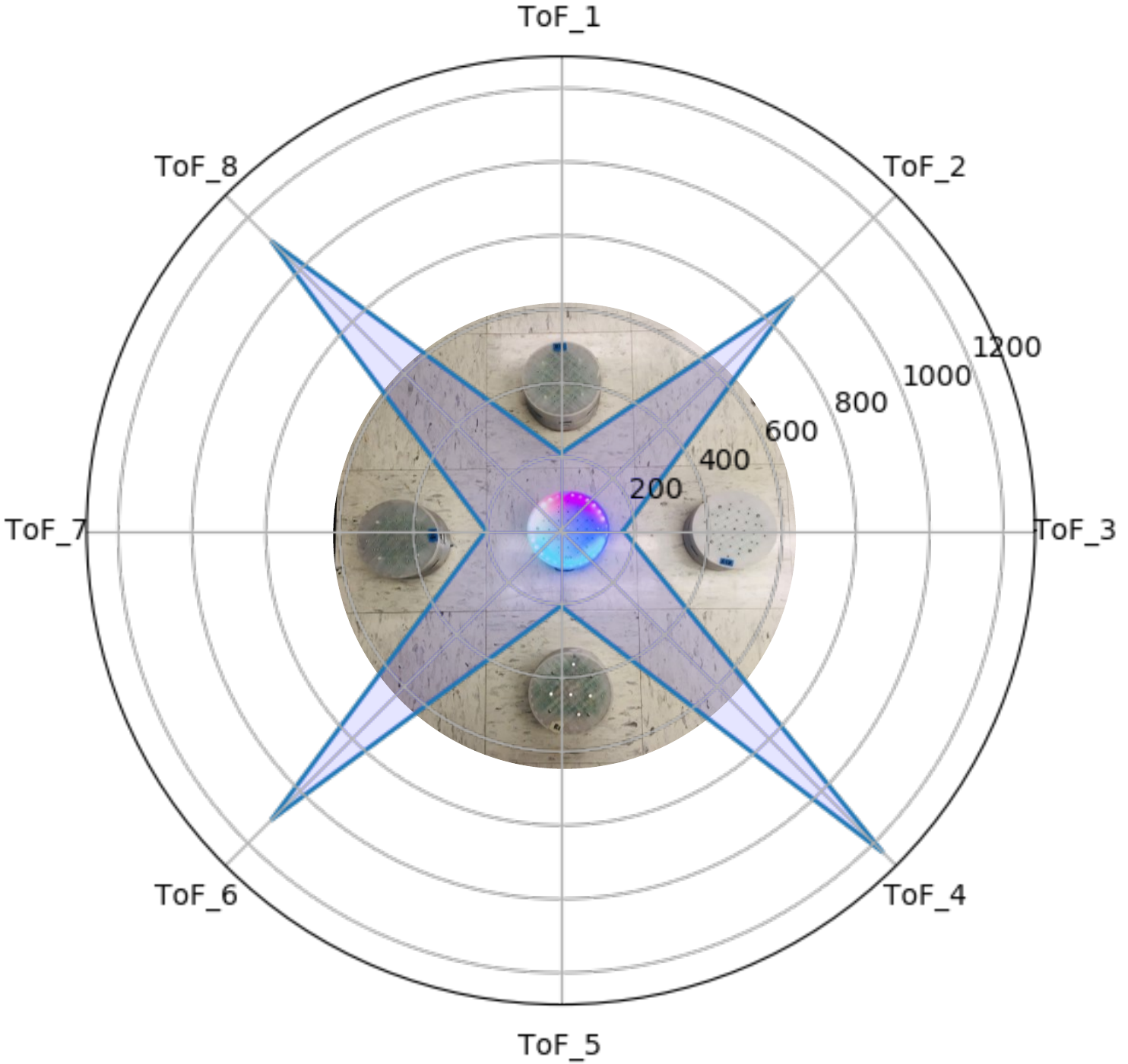}
        \caption{}
        \label{img:sensor_exp_1}
    \end{subfigure}
    \begin{subfigure}[b]{0.22\linewidth}
        \centering
        \includegraphics[width=1\linewidth]{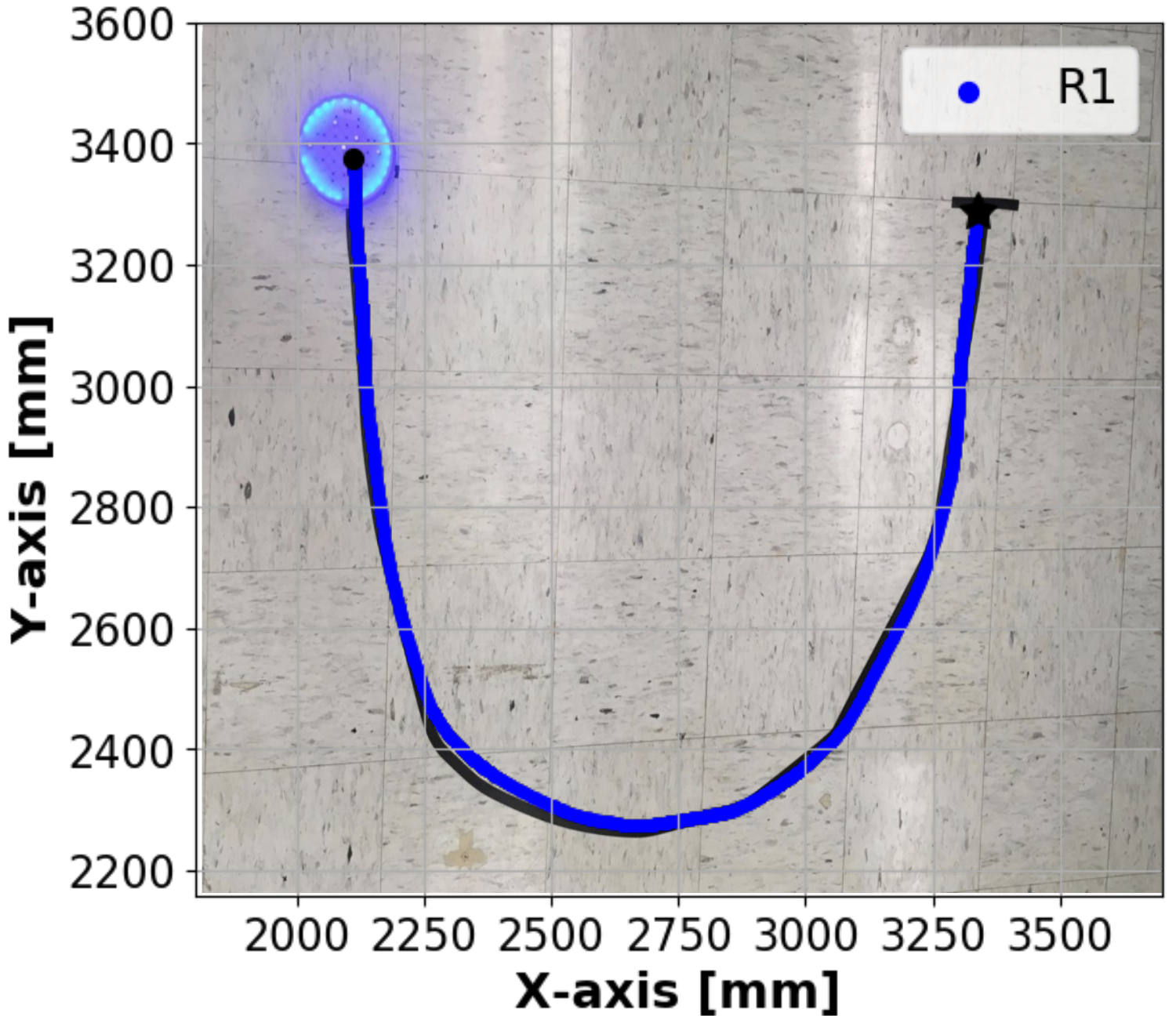}
        \caption{}
        \label{img:sensor_exp_2}
    \end{subfigure}
    \begin{subfigure}[b]{0.28\linewidth}
        \centering
        \includegraphics[width=1\linewidth]{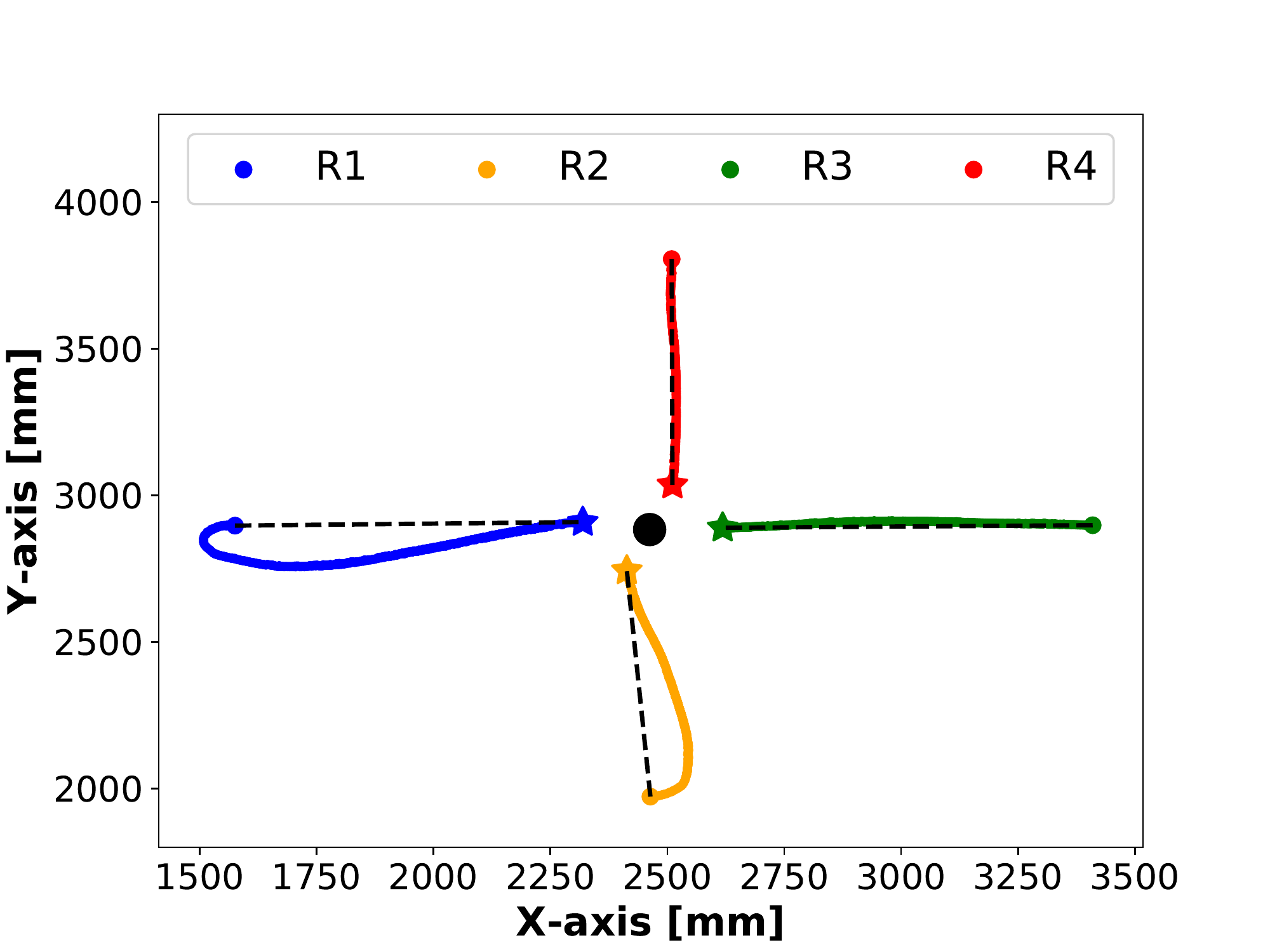}
        \caption{}
        \label{img:go_to_goal_exp}
    \end{subfigure}
    \begin{subfigure}[b]{0.28\linewidth}
        \centering
        \includegraphics[width=1\linewidth]{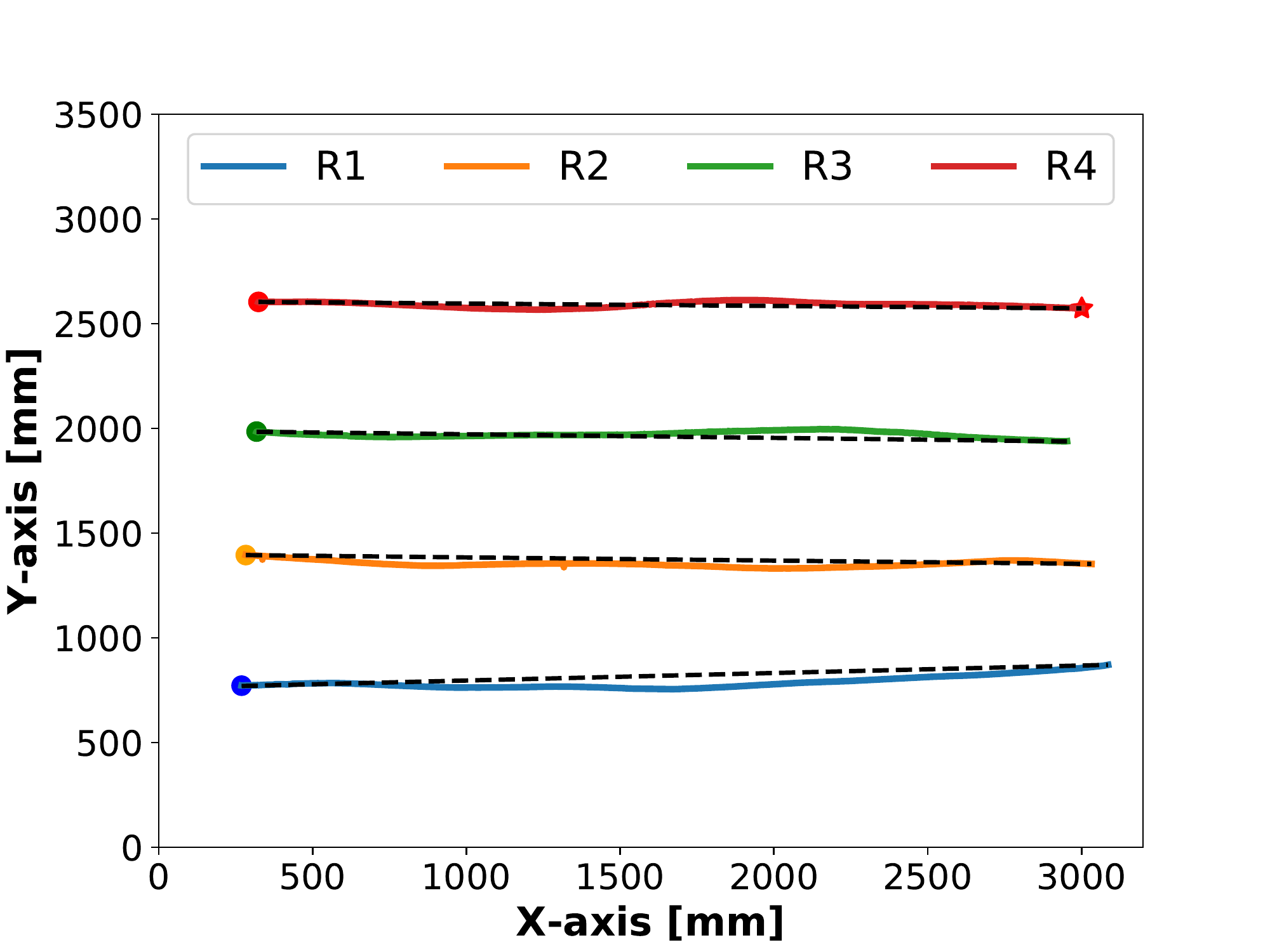}
        \caption{}
        \label{img:pure-pursuit_exp}
    \end{subfigure}

\caption{Results of the function and performance tests on the SMARTmBOT: (a) the ToF sensors test, (b) the line tracing test using only two line sensors, (c) the go-to-goal navigation algorithm test, and (d) the pure-pursuit navigation algorithm test. In order to validate the reproducibility of the proposed platform, multiple robots were manufactured using the same manufacturing process, and two navigation algorithm tests (c) and (d) were conducted.}   
    \label{img:exp_result}
\end{figure*}

As a result, all four robots could follow the given waypoints well and arrive at their respective goal positions using this algorithm. Their trajectories are illustrated in Fig. \ref{img:pure-pursuit_exp}, where the colored dots indicate the starting positions of each robot.

\section{Potential Applications}
\label{sec:application}
\begin{figure*}[t]
    \centering
    \begin{subfigure}[b]{0.169\linewidth}
        \centering
        \includegraphics[width=1\linewidth]{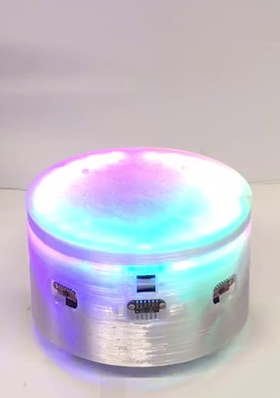}
        \caption{} 
        \label{img:robot_app_rgb}
    \end{subfigure}
    \begin{subfigure}[b]{0.1515\linewidth}
        \centering
        \includegraphics[width=1\linewidth]{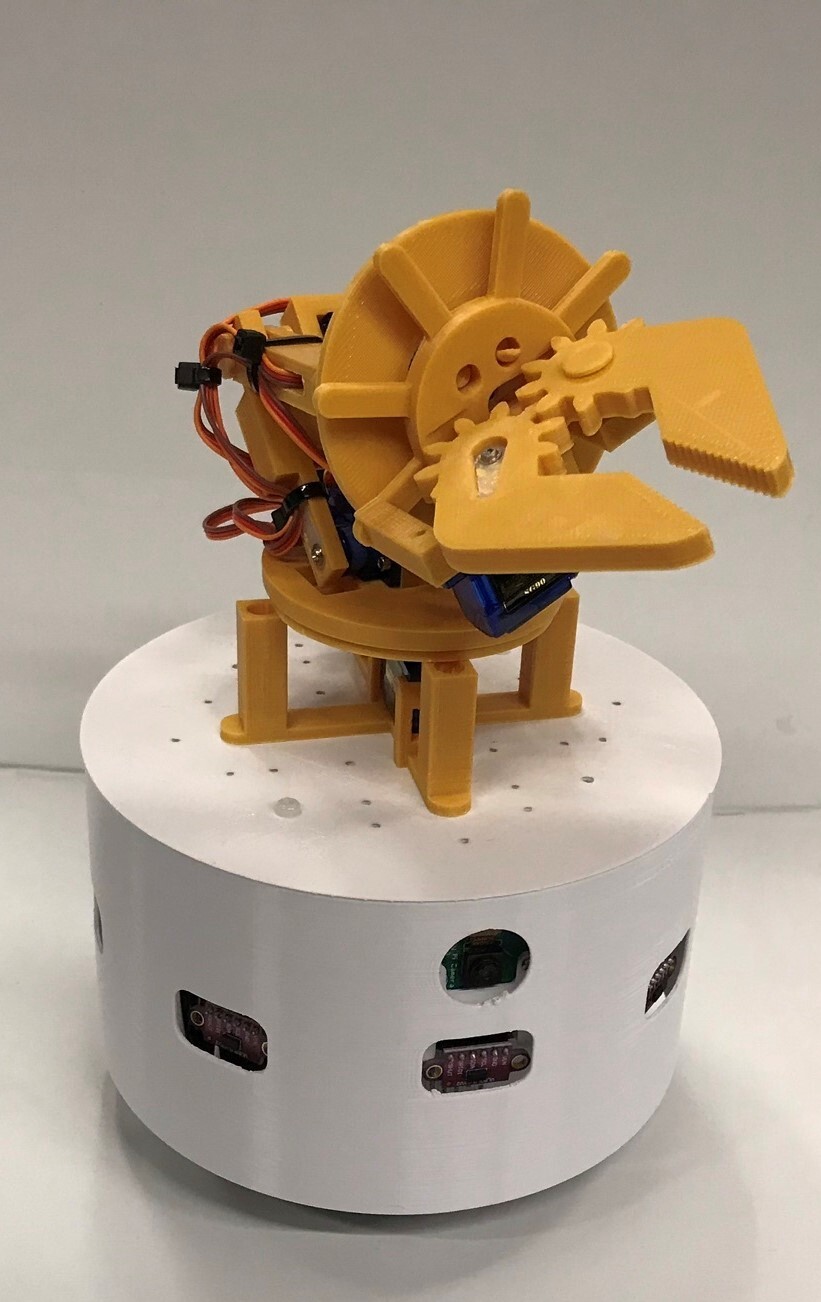}
        \caption{} 
        \label{img:robot_app_arm}
    \end{subfigure}
    \begin{subfigure}[b]{0.32\linewidth}
        \centering
        \includegraphics[width=1\linewidth]{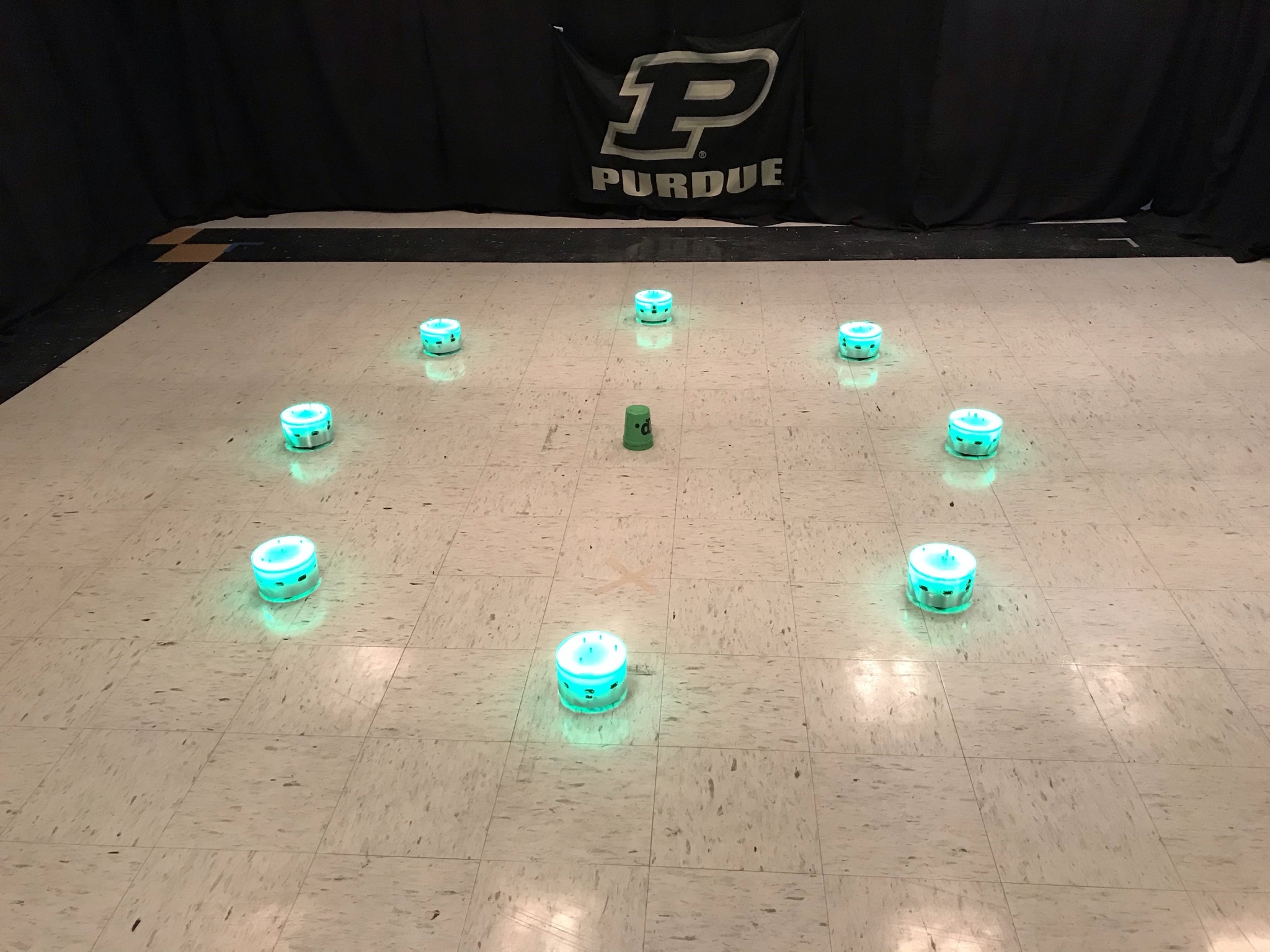}
        \caption{} 
        \label{img:robot_app_swarm}
    \end{subfigure}
    \begin{subfigure}[b]{0.3185\linewidth}
        \centering
        \includegraphics[width=1\linewidth]{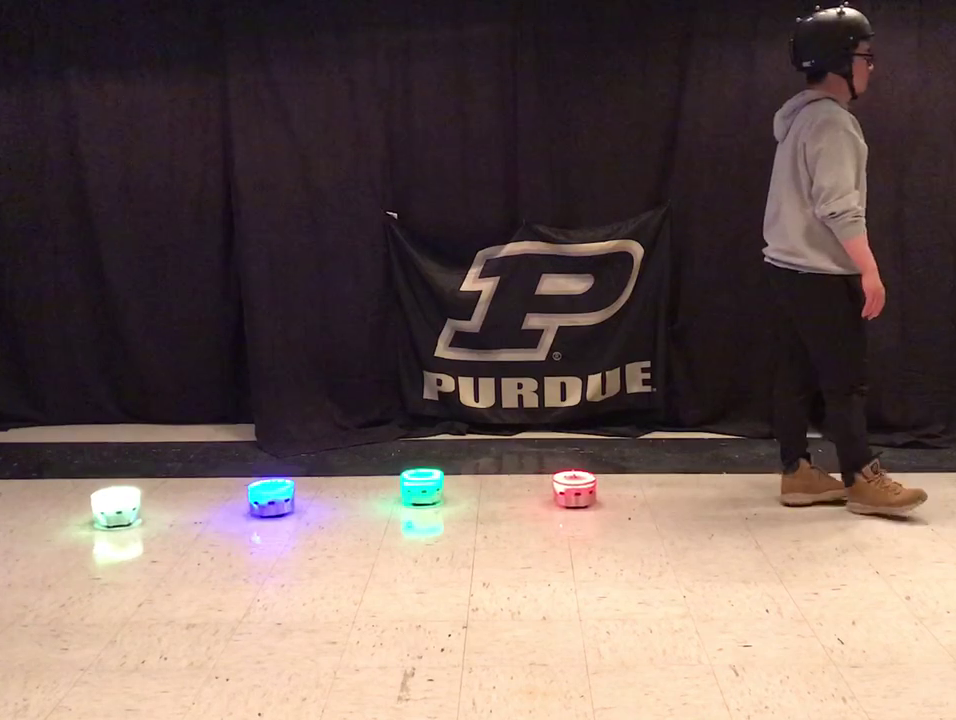}  
        \caption{} 
        \label{img:robot_app_hri}
    \end{subfigure}
    \caption{Potential applications using the SMARTmBOT: (a) a color changing SMARTmBOT case using RGB LED strips, (b) a miniature robot arm mounted on the SMARTmBOT, (c) an example of the rendezvous algorithm as swarm research, and (d) an example of the leader-follower algorithm as HRI research. A demo video that shows each of these applications is available in the supplementary video.
    }
    \label{img:robot_app}
\end{figure*}

To demonstrate the expandability and applicability of the SMARTmBOT in various respects, this section introduces potential applications for connecting it with other platforms (e.g., RGB LED strips and a robot arm) and utilizing the platform for swarm and human-robot interaction (HRI) research, illustrated in Fig. \ref{img:robot_app}. 

\subsection{Adjustable Color LED Case using RGB LED Strip}
Fig. \ref{img:robot_app_rgb} shows an example of the SMARTmBOT platforms connected to a flexible RGB LED strip capable of changing the color of the case. The RGB strip was connected to the extra pins provided on the third PCB layer and is controlled through an additional ROS2 node named \textit{Writing\_WB2813b\_RGB\_STRIP}, with the topic type \textit{std\_msgs/msg/Float32MultiArray} in the SMARTmBOT package. The node can change the colors, intensity of illumination, the number, and the motions of the LEDs (e.g., chasing or rainbow effect). The proposed LED case can be utilized for visualizing the selected robots, displaying the current status of the robots, among other uses.

\subsection{Mobile Manipulators Robots Using 5--DoF Robot Arm}
Fig. \ref{img:robot_app_arm} shows an example of combination with other robotic platforms. For example, we mounted an open-source 5--DoF robot arm platform, which is small and composed of five RC servo motors, on the top of the SMARTmBOT and added an I2C-based PWM extension board (e.g., PCA9685 16 channel 12 Bit PWM servo driver board) with an extra battery for the servo motors. The extension board is connected to the extra I2C pins on the third PCB layer. The robot arm platform was attached to the top of the SMARTmBOT via M3 holes of the case. Similar devices can be added using the same method.

\subsection{Swarm Controller Research}
Fig. \ref{img:robot_app_swarm} depicts an example of utilizing multiple SMARTmBOTs as swarm research, such as multi-agent rendezvous control. Rendezvous control is one of the swarm controllers for all agents to converge on a goal position \cite{bresson2017simultaneous}. In the experiment, position and orientation of each SMARTmBOT was tracked through the motion capture system, and the simple rendezvous controller was utilized to have eight SMARTmBOTs gather at their goal positions (e.g., the green cone). To easily identify the current status of each robots, we changed the case color via the RGB strip node (green indicates that it is still not arriving at the goal position). Based on this experiment, we expect SMARTmBOTs to combine with other similar multi-agent control systems fluidly.

\subsection{Human-Robot Interaction Research}
Fig. \ref{img:robot_app_hri} depicts an example of using multiple SMARTmBOTs as HRI research. The goal of this experiment is to engage a leader-following algorithm \cite{consolini2008leader} such that a swarm team consisting of four SMARTmBOTs follows a human leader by maintaining a line and gaps between the human leader or other robots. In the experiment, the human leader wore a helmet with reflective markers attached. The robot positions and orientations were tracked via a motion capture system. The experiment displayed SMARTmBot's potential for investigating human-robot interaction.

\section{Conclusion and Future works}
\label{sec:conclusion}

In this paper, we presented a ROS2-based, low-cost, open-source mobile robot platform that can be applied to research endeavors in various robotic disciplines on account of its modular-typed and customizable extensions. The SMARTmBOT contains adjustable 3D-printed parts and replaceable sections that allow ready customization of robot functionality for greater flexibility of use. In addition, the system has been tested for both mobility and sensor integration, showing that the base platform works as a standalone robot.

This platform has the capability to be extended to applications in a great number of fields, including both research and education. With the listed components, adjustments are simple to make by repurchasing or 3D-printing the necessary parts, making the robot adaptable to many contexts within robot control and sensor usage. Also, copies of the base robot can be made, providing a multi-robot platform for swarm robotics applications. 

In the future, we will continue to update this platform to work with future ROS2 version updates and identify additional use cases for the robot. We plan to share the mobile platform with K-12 students to give them additional opportunities to learn about human-robot interactions, and to broaden the horizons of robotics with the endless possibilities afforded by a cheap and flexible open-source mobile robot platform.

\section*{Online Repositories}
The GitHub repository can be found in the following link: \url{https://github.com/SMARTlab-Purdue/SMARTmBOT}. In the repository, there are 3D design CAD files, PCB design files, ROS2 package utilized on SMARTmBOT, and simple robot navigation controllers (e.g., go-to-goal, pure-pursuit, and line tracing codes). The 3D design files are created with SolidWorks 2020, the PCB designs are created with AutoDesk Eagle CAD, and the ROS2 nodes are written using Python 3.8.

\section*{ACKNOWLEDGMENT}
This material is based upon work supported by the National Science Foundation under Grant No. IIS-1846221. Any opinions, findings, and conclusions or recommendations expressed in this material are those of the author(s) and do not necessarily reflect the views of the National Science Foundation.

\bibliography{root}
\bibliographystyle{IEEEtran}

\end{document}